\documentclass[conference]{IEEEtran}
\IEEEoverridecommandlockouts
\usepackage{cite}
\usepackage{amsmath}
\usepackage{amssymb}
\usepackage{amsfonts}
\usepackage{algorithmic}
\usepackage{graphicx}
\usepackage{textcomp}
\usepackage{xcolor}
\usepackage{booktabs}
\usepackage{epsfig}
\usepackage{hyperref}
\usepackage{pgfplots}
\usepackage{import}
\usepackage{ulem}
\usepackage{subfigure}
\usepackage{booktabs}
\usepackage{makecell}
    
\usepackage{siunitx}
\pgfplotsset{width=7cm,compat=1.9}

\usepackage{wrapfig}
\usepackage{booktabs}
\usepackage{multirow,tabularx,capt-of}
\def\BibTeX{{\rm B\kern-.05em{\sc i\kern-.025em b}\kern-.08em
    T\kern-.1667em\lower.7ex\hbox{E}\kern-.125emX}}

\graphicspath{ {./images/} }

\begin{document}

\title{Comparative Study of Parameter Selection for Enhanced Edge Inference for a Multi-Output Regression model for Head Pose Estimation}
\author{
\IEEEauthorblockN{Asiri Lindamulage, Nuwan Kodagoda, Shyam Reyal, Pradeepa Samarasinghe} 
\IEEEauthorblockA{\textit{Faculty of Computing,} \\
\textit{Sri Lanka Institute of}\\
\textit{ Information Technology}\\
Sri Lanka.\\
asiri.l, nuwan.k, shyam.r, pradeepa.s@sliit.lk}
\and
\IEEEauthorblockN{Pratheepan Yogarajah} 
\IEEEauthorblockA{\textit{School of Computing} \\
\textit{Engineering and Intelligent Systems,}\\
\textit{Ulster University}\\
United Kingdom.\\
p.yogarajah@ulster.ac.uk}
}
\maketitle
\begin{abstract}
Magnitude-based pruning is a technique used to optimise deep learning models for edge inference. We have achieved over 75\% model size reduction with a higher accuracy than the original multi-output regression model for head-pose estimation.
\end{abstract}

\begin{IEEEkeywords}
Edge Inference, Optimisation, Network Pruning, Quantisation, TensorFlow,Head Pose estimation
\end{IEEEkeywords}


\section{Introduction}
With the development of deep learning models and the enhancement of technology used in edge devices such as mobile phones and tablets, the deployment of deep learning models in edge devices has become popular \cite{neuralNetworkonSoc}. Being able to run a deep learning inference without an internet connection or a connection to a server is a huge advantage to save network bandwidth\cite{EnergyAwareEdgeInference}, energy and to reduce network latency related delays considerably \cite{AIBenchmarkAnroid}. Furthermore, with edge inference, user data and the data used in the inference cycle do not leave the device, providing users with a secure and enhanced user experience\cite{MlPerf}.

In order to deploy a deep learning model, the model should be able to work with the limited amount of resources available on the edge device\cite{AutomatedEnsamble}. A mobile device cannot handle a heavy model that is intended to be deployed on a server-level computer. Hence, those should be optimised. Network pruning, quantisation, parameter factorisation, tensor decomposition, and teacher-student training \cite{Empirical} are some of the Deep Neural Network optimisation (DNN) techniques. Furthermore, there are mobile-friendly DNN architectures such as MobileNet\cite{mobileNet}, EfficientNet\cite{efficientNet}, SqueezeNet\cite{squeezenet} etc., which are optimised to be deployed on edge devices due to their lower number of parameters and lower model sizes but get the same level or even better accuracy than non-optimised models.

Taylor approximation and Hessian approximation, inverse Hessian approximation, and magnitude-based pruning have been used for model pruning since 1990s\cite{Prune_not_to_prune}. Magnitude-based pruning has become popular since it can optimise large networks regardless of the network architecture\cite{Prune_not_to_prune,Liang2021a}. Michael H. Zhu and Suyog Gupta have presented magnitude-based model pruning as a module in the TensorFlow Model Optimisation library\cite{Prune_not_to_prune}. Although the impact of magnitude-based model pruning on the model size and model accuracy on classification models has been discussed before \cite{Prune_not_to_prune}, the impact on a regression model has not been analysed.

Our work focuses on an in-depth analysis of the impact of magnitude-based pruning on the regression models and how the pruning parameters have to be tuned to obtain a model with a smaller model size while preserving model accuracy. Therefore, this study seeks to find out how a regression-based model can be compressed by tuning the pruning parameters without changing any hyper-parameters or architecture of the model. Our key contributions of this study are:
\begin{itemize}
\item Conduct in-depth analysis on parameter selection for the model pruning to get the optimal solution for the best model size while preserving the model accuracy.
\item Compare the impact of Constant and Dynamic pruning.
\end{itemize}

\section{Methodology}
This section describes the types of magnitude-based pruning, the construction of the model, and the testing criteria for measuring performance and the model size of the resulting model when magnitude-based pruning is applied to the multi-output regression model to optimise the model for edge inference.

Magnitude-based pruning makes the weights zero if they are less than a given threshold. By making weights zero, the computational power and propagation delays required when making a prediction can be reduced since the weights that are zero can be eliminated in the computations. Based on how this threshold is chosen and how the sparsity of the model is increased in the training process, there are two methods of model pruning: dynamic pruning and constant pruning.
\subsection{Dynamic pruning and Constant pruning}

Dynamic pruning allows one to gradually increase the sparsity percentage. The equation of dynamic pruning is as follows: \cite{Prune_not_to_prune}
\[s{_t}=s{_f}+(s{_i}-s{_f})\left(1-\frac{t-t{_0}}{(t{_f}-t{_0})\Delta t}\right)^3,\]
where\\
$s{_t}$\quad = Current sparsity value\hspace{10mm}$s{_i}$\quad = Initial sparsity
\\$s{_f}$\quad = final sparsity\hspace{10mm}$t{_0}$\quad = Starting training step
\\$t{_f}$\quad = Ending training step\hspace{10mm}$\Delta t$\quad = Pruning frequency
\\
The layers chosen to be pruned are assigned a binary mask variable of the shape and size of the weights of the layer, which is used to determine which weights participate in the forward propagation. In each epoch, the weights are sorted by the absolute value and the smallest weights' masks are changed to zero. In the back propagation process, the weights masked with zeros are not updated. Once the desired sparsity is achieved, the mask updating process will be stopped. This process is continued until the desired sparsity value $s{_f}$ is reached. By controlling the above parameters, we can fine-tune the accuracy and size of the generated pruned model.

But in constant pruning, a constant sparsity level $s{_c}$ is maintained throughout the training process. When the pruning process starts, it immediately tries to reach the desired sparsity level. This sparsity value doesn't depend on the starting step, end step, and frequency of constant pruning, but they have to be tuned to get the best outcome of the constant pruning method.

\subsection{Model}
\begin{figure}[t]
    \centering
    \includegraphics[width=10cm]{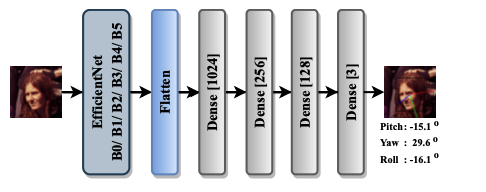}
    \caption{Model Architecture (Source:-\cite{Dhanawansa})}
    \label{fig:model}
\end{figure}

The model chosen for the experiments of this study is a multi-output regression model by Dhanawnsa.V et al. \cite{Dhanawansa} which was implemented for head-pose estimation. It is a custom model based on the Efficient B0 model\cite{efficientNet} combined with additional dense layers at the top of the architecture. The hyperparameters used in the training process are as follows:
\begin{itemize}
    \item Learning rate = 0.001
    \item Optimiser = Adam
    \item Batch size = 128
    \item No of total training cycles(epochs) = 80
    \item Loss Function = Mean Absolute Error (MAE)
    
    \item Learning rate decay function = ReduceLROnPlateau
    \begin{itemize}
             \item patience = 3
             \item factor = 0.5
             \item min learning rate = 0.00001
         \end{itemize}
    \item Data-set = BIWI Kinect Head Pose \cite{fanelli_IJCV}, 300W-LP \cite{300WLP} and AFLW 2000 \cite{300WLP}     
\end{itemize}
The input to the model is a RGB image of 64$\times$64 and the output provides yaw, pitch and roll angles for head-pose estimation\cite{Dhanawansa} (see Fig. \ref{fig:model}). All parameters mentioned above were kept constant through all experiments conducted in this study.

\subsection{Model Pruning implementation }
Implementation was done to test the performance of constant pruning and dynamic pruning, varying one parameter at a time. When model pruning is applied, some of the layers have to be neglected since they are not supported by the model pruning library. Those layers are \textit{re-scaling layers, normalisation layers, and multiplication layers.} They were annotated not to be included in the pruning process.

The pruning parameters which will be analyzed through this study are as follows,
\begin{enumerate}
     \item Pruning schedule
     \begin{itemize}
         \item Constant pruning - $s{_c}$
         \item Dynamic pruning
         \begin{itemize}
             \item Initial sparsity - $s{_i}$
             \item Final sparsity - $s{_f}$
         \end{itemize}
     \end{itemize}
     \item Starting training step (epoch) - $t{_0}$
     \item Ending training step (epoch) - $t{_f}$
\end{enumerate}

We experimented with 50\%, 75\%, and 87.5\% final sparsity ($s_f$) values for both dynamic and constant pruning because values less than 50\% result in weights less than 50\%, which does not result in any significant improvement over the model size, and values greater than 87.5\% completely destroy the model accuracy because all significant weights that contain important features are also eliminated. Starting step values ($t_0$) were 0, 20, 40, 60, 80, and ending step values ($t_f$) were combinations of 20, 40, 60, 80.

\subsection{Testing criteria}
To maximise the impact of model pruning, Gzip compression was used\cite{Prune_not_to_prune}. Although model pruning makes weights zero based on the desired sparsity value, those zero weights are still represented as a 32-bit number, which takes up the same amount of space as none-zero weights. Therefore, to get the maximum benefit of model pruning, those 0 values have to be removed by compressing the file \cite{TensorFlow2021}. The model accuracy was measured based on the Mean Absolute Error (MAE) as per the original model. \textbf{The accuracy of the base model used in this study is 8.63 with a file size of 110MB} and all tests are compared with the base model accuracy.

For each test, 3 models were implemented. The models discussed through this study are TensorFlow Lite models which are ready to be deployed on edge devices. 
\begin{enumerate}
    \item Pruned model - Without any post optimiser on model size and accuracy .
    \item Optimised model - Used ``EXPERIMENTAL\_SPARSITY'' post-optimiser on model size\cite{tfliteopt}.
    \item Quantized model - Used ``DEFAULT'' post-optimiser to quantise(32 bit to 8 bit) the pruned model \cite{tfliteopt} 
\end{enumerate}
\label{section:testingcr}

\section{Results and discussion}

\subsection{Dynamic Sparsity}

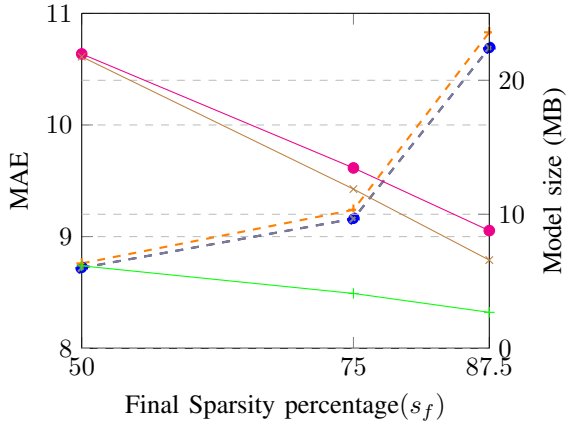
\begin{figure}
\centering
\begin{tikzpicture}

\begin{axis}[
    legend entries={Pruned model MAE,Optimized model MAE ,Quantized model MAE},
    legend to name=all_maes,
    xlabel={Final Sparsity percentage$(s{_f})$},
    ylabel={MAE},
    xmin=50, xmax=87.5,
    ymin=8, ymax=11,
    xtick={50,75,87.5,100},
    ymajorgrids=true,
    grid style=dashed,
    axis y line* = left, 
]
\addplot[
    dashed, 
    thick,
    color=blue,
        mark=*]
    coordinates {
    (50, 8.72)
    (75,9.16)
    (87.5,10.69)
    };
\addplot[
          dashed, 
    thick,
     color=gray,
        mark=x]
    coordinates {
    (50, 8.72)
    (75,9.16)
    (87.5,10.69)
    };
\addplot[
          dashed, 
    thick,
    color=orange,
        mark=+]
    coordinates {
    (50,8.76)
    (75,9.24)
    (87.5,10.83)
    };
\end{axis}

\begin{axis}[
    legend entries={Pruned model Size,Optimized model Size ,Quantified model Size},
    legend to name=sizes_alls,
    ylabel={Model size (MB)},
    xmin=50, xmax=87.5,
    ymin=0, ymax=25,
    xtick={50,75,87.5,100},
    ymajorgrids=true,
    grid style=dashed,
            hide x axis,
        axis y line*=right,
]

\addplot[
    color=magenta,
    mark=*
    ]
    coordinates {
    (50,21.97)
    (75,13.46)
    (87.5,8.78)
    };
\addplot[
    color=brown,
    mark=x
    ]
    coordinates {
    (50,21.76)
    (75,11.88)
    (87.5, 6.58)
    };
\addplot[
    color=green,
    mark=+
    ]
    coordinates {
    (50,6.15)
    (75,4.09)
    (87.5, 2.68)
    };

\end{axis}
\end{tikzpicture}

\caption{Final Sparsity percentage $(s{_f})$ vs model size and MAE}
\label{poly1}
\end{figure}

Fig. \ref{poly1} shows how the model size (in MB) and the Mean Absolute Error (MAE) vary when the final sparsity ($s{_f}$) is increased from 50\% up to 87.5\%. The MAE value has increased, which means the accuracy of the model has decreased when the final sparsity ($s{_f}$) is increased. But the file size has decreased considerably. The file size has been reduced since the number of zero weights increases with sparsity. With the model becoming sparse, it might lose weights that contain important features that contribute to the accuracy of the model. Therefore, there's a trade-off between model size and model accuracy when pruning is applied to a model. Since the MAE curves of the pruned model and post-pruning optimised model overlaps with each other, we can observe that by applying the post-pruning optimiser, the model size could be slightly reduced further without making any difference in the model accuracy. Furthermore, when post quantisation is applied to the model, the model size has been reduced significantly more than the other two models, with a very small reduction in model accuracy. By applying only pruning, the size of the model could be reduced by up to 5.9\% (8.78MB) - 19.7\% (21.97MB) of the original model size (110MB). By applying the post-pruning optimiser, the model size could be further reduced by up to 2.2MB of the pruned model. Quantifying the pruned model could further reduce the model size up to 15.82MB of the pruned model, which is 5.59\% (6.15MB) - 2.43\% (2.68MB) of the original model size.

\begin{figure}
\ref{all_maes}
\ref{sizes_alls}
\end{figure}

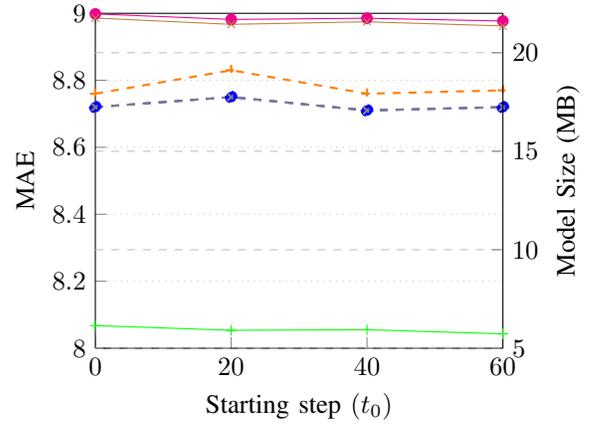
\begin{figure}
\centering
\begin{tikzpicture}[]
\begin{axis}[
    xlabel={Starting step $(t{_0})$},
    ylabel={MAE},
    xmin=0, xmax=60,
    ymin=8, ymax=9,
    xtick={0,20,40,60},
    ymajorgrids=true,
    grid style=dotted,
    axis y line* = left,
]

\addplot[
dashed, 
    thick,
    color=blue,
    mark=*
 ]
    coordinates {
    (0, 8.72)
    (20,8.75)
    (40,8.71)
    (60,8.72)
    };
 
\addplot[
dashed, 
    thick,
    color=gray,
    mark=x
 ]
    coordinates {
    (0, 8.72)
    (20,8.75)
    (40,8.71)
    (60,8.72)
    };
    
\addplot[
dashed, 
    thick,
    color=orange,
    mark=+
 ]
    coordinates {
    (0, 8.76)
    (20,8.83)
    (40,8.76)
    (60,8.77)
    };

\end{axis}
\begin{axis}[
    xlabel={Starting step \%},
    ylabel={Model Size (MB)},
    xmin=0, xmax=60,
    ymin=5, ymax=22,
    xtick={0,20,40,60},
    ymajorgrids=true,
    grid style=dashed,
    hide x axis,
    axis y line*=right,
]

\addplot[
    color=magenta,
    mark=*
    ]
    coordinates {
    (0, 21.97)
    (20,21.69)
    (40,21.75)
    (60,21.61)
    };
\addplot[
    color=brown,
     mark=x
    ]
    coordinates {
    (0,21.76)
    (20,21.44)
    (40,21.57)
    (60,21.36)
    };
\addplot[
    color=green,
     mark=+
    ]
    coordinates {
    (0,6.15)
    (20,5.91)
    (40,5.94)
    (60,5.73)
    };
\end{axis}
\end{tikzpicture}
    \caption{Starting step  $(t{_0})$ vs model size and MAE (0-50\% Dynamic Sparsity, 80 - end step $(t{_f})$)}
    \label{sparsityPoly_50}
\end{figure}

\begin{figure}
\centering
\begin{tikzpicture}[]
\begin{axis}[
    xlabel={Ending step $(t{_f})$},
    ylabel={MAE},
    xmin=20, xmax=80,
    ymin=8, ymax=9,
    xtick={20,40,60,80},
    ymajorgrids=true,
    grid style=dotted,
    axis y line* = left,
]

\addplot[
    dashed, 
    thick,
    color=blue,
    mark=*
 ]
    coordinates {
    (20, 8.65)
    (40,8.65)
    (60,8.54)
    (80,8.72)
    };
 
\addplot[
    dashed, 
    thick,
      color=gray,
    mark=x
 ]
    coordinates {
    (20, 8.65)
    (40,8.65)
    (60,8.54)
    (80,8.72)
    };
 
\addplot[
    dashed, 
    thick,
    color=orange,
    mark=+
 ]
    coordinates {
    (20, 8.68)
    (40,8.72)
    (60,8.58)
    (80,8.76)
    };

\end{axis}
\begin{axis}[
    xlabel={Ending step \%},
    ylabel={Model Size (MB)},
    xmin=20, xmax=80,
    ymin=5, ymax=23,
    xtick={20,40,60,80},
    ymajorgrids=true,
    grid style=dashed,
    hide x axis,
    axis y line*=right,
]

\addplot[
    color=magenta,
    mark=*
    ]
    coordinates {
    (20, 22.32)
    (40,22.17)
    (60,22.08)
    (80,21.97)
    };
\addplot[
    color=brown,
     mark=x
    ]
    coordinates {
    (20,22.02)
    (40,21.90)
    (60,21.82)
    (80,21.76)
    };
\addplot[
    color=green,
     mark=+
    ]
    coordinates {
    (20, 6.17)
    (40,6.12)
    (60,6.24)
    (80,6.15)
    };
\end{axis}
\end{tikzpicture}
    \caption{Ending step $(t{_f})$ vs model size and MAE (0-50\% Dynamic Sparsity, 0 - starting step$(t{_0})$)}
    \label{sparsityPoly_end}
\end{figure}
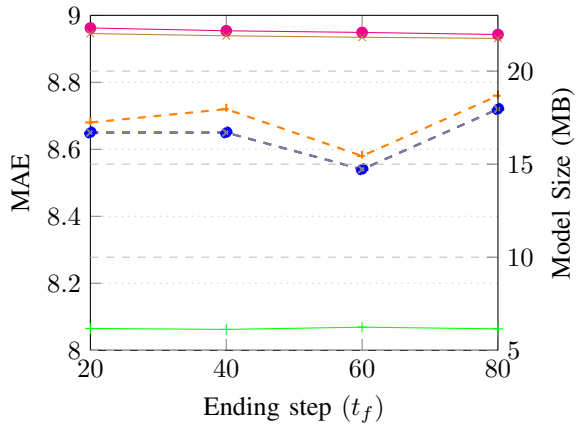

According to Fig. \ref{sparsityPoly_50} the best model accuracy was obtained when pruning was started at the 40$\textsuperscript{th}$ epoch. The model sizes are almost constant throughout the starting steps ($t{_0}$) on all sparsity levels, but the accuracy keeps going up and down in a small range. Similarly, the Fig. \ref{sparsityPoly_end} shows the model size and MAE when the ending step ($t{_f}$) is increased from 0 to 80, with $t{_0}=0$ and $s{_f} = 50\%$. Observing the graph, the best accuracy level is reached when the ending step ($t{_0}$) is 60. This indicates that the model can recover the loss that is caused by the pruning process while maintaining the same level of model size if it has enough training epochs after the desired sparsity level is reached. But the number of pruning epochs cannot be too low either. The desired sparsity level has to be reached gradually without harming model accuracy. It can be observed that when stopped at the 80$\textsuperscript{th}$ epoch, the model has lost accuracy since it did not have enough time to recover the lost accuracy. Observing the experimental results, the starting step and the ending step cannot be considered separately. The combination of both is what matters to the accuracy of the pruned model. Overall, the model with the best accuracy was obtained at $t{_0} = 0$ and $t{_f} = 60$.

Furthermore, tests were run to see how initial sparsity ($s_i$) affected model size and accuracy, but there was no significant difference. MAE varied in the range between 8.9 and 9.96 throughout the tests. Overall, the best accuracy in dynamic pruning was achieved at 50\% final sparsity ($s{_f}$) when $t{_0}$ = 0 and $t{_f}$ = 60, with a total number of 80 training steps and applying the post-pruning optimiser. The resulting model file size was 21.82MB (19.83\% of the original model size) with an MAE value of 8.54 (0.09 lower than the original model), which is slightly better than the original model. The model with the best file size was obtained at $s{_f}$ = 87.5\%, $t{_0}$ = 40, $t{_f}$ = 80 and the post-pruning optimiser was applied. The model accuracy (MAE) is 11.99 (-3.36 worse than the original model) with a file size of 8.56 (7.78\% of the original model size). By applying post quantisation, the best accuracy model size could be further reduced to 6.24MB (5.67\% of the original model size) with an 8.58 MAE value (0.05 lower than the original model), which is still a bit better than the original model. The model that had the best file size could be further reduced to 2.54MB (2.3\% of the original model size) with a further reduced accuracy of 12.17 (MAE).

\subsection{Constant Sparsity}

\begin{figure}
\centering
\begin{tikzpicture}

\begin{axis}[
    xlabel={Static sparsity percentage \% $(s{_f})$},
    ylabel={MAE},
    xmin=50, xmax=87.5,
    ymin=8, ymax=11,
    xtick={50,75,87.5,100},
    ymajorgrids=true,
    grid style=dotted,
    axis y line* = left, 
]

\addplot[
    dashed, 
    thick,
    color=blue,
    mark=*
 ]
    coordinates {
    (50,8.58)
(75,9.27)
(87.5,10.04)
    };

\addplot[
    dashed, 
    thick,
     color=gray,
    mark=x
 ]
    coordinates {
    (50,8.58)
(75,9.27)
(87.5,10.04)
    };
 
\addplot[
    dashed, 
    thick,
    color=orange,
     mark=+
 ]
    coordinates {
    (50,8.61)

(75,9.30)
(87.5,10.12)
    };

\end{axis}
\begin{axis}[
    ylabel={Model size (MB)},
    xmin=50, xmax=87.5,
    ymin=0, ymax=25,
    xtick={50,75,87.5,100},
    ymajorgrids=true,
    grid style=dashed,
            hide x axis,
        axis y line*=right,
]

\addplot[
    color=magenta,
    mark=*
    ]
    coordinates {
(50,22.29)
(75,13.83)
(87.5,9.08)
    };
\addplot[
    color=brown,
    mark=x
    ]
    coordinates {
(50,21.98)
(75,12.13)
(87.5, 6.72)
    };
\addplot[
    color=green,
    mark=+
    ]
    coordinates {
(50,6.21)
(75,4.32)
(87.5, 2.90)
    };

\end{axis}

\end{tikzpicture}
    \caption{Static sparsity $(s{_f})$ vs model size and MAE}
    \label{sparsity1}
\end{figure}
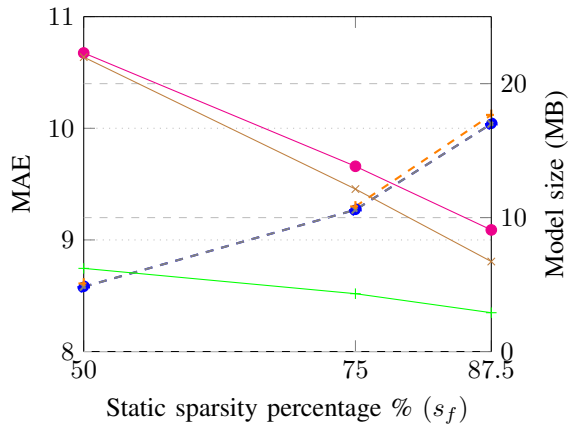

Inspecting the Fig. \ref{sparsity1}, the accuracy and model size behaviour with sparsity level is the same as dynamic sparsity, but the range of accuracy differs from the original model (0.02 - 1.68) and the file size range of the resulting models (8.25\% - 20.26\% of the original model) is also higher than dynamic pruning. The behaviour of the models generated by applying the quantisation and post-pruning flags is also similar to that of dynamic pruning. The effect of the generated model's starting step ($t_0$) and ending step ($t_f$) values, as well as the variation in model size and accuracy, was the same as that of dynamic pruning.

Overall, the best model accuracy of 8.57 (MAE) is obtained with 50\% sparsity ($s{_c}$) and the pruning process starting ($t{_0}$) from the 60$\textsuperscript{th}$ to 80$\textsuperscript{th}$ epoch($t{_f}$), which is 0.06 better than the original model with a relatively larger file size of 35.54MB (32.3\% of original model). The model with the best file size (8.41) was obtained at 0.88 sparsity with the pruning process starting from the 60$\textsuperscript{th}$ to 80$\textsuperscript{th}$ epoch with a reduced accuracy of 12.39 (3.76 lower than the original model). The best-accurate model size could be reduced to 8.59MB (7.8\% of the original model size) with an accuracy of 8.59 (0.02 lower than the best model) by applying post quantisation.

\subsection{Constant pruning Vs Dynamic Pruning}

\begin{table*}[t]
    \centering
\caption{Best models of Dynamic and Constant pruning}
\begin{tabular}{l c c c c c c c c c c} 
\toprule
\thead{Model} & \multicolumn{4}{c}{\thead{Training parameters}} & \multicolumn{2}{c}{\thead{Pruned model}} & \multicolumn{2}{c}{\thead{Post pruned model}} & \multicolumn{2}{c}{\thead{Post quantized model}} \\ 
\cmidrule{1-11}
& \thead{Initial\\Sparsity} & \thead{Final\\Sparsity} & \thead{Starting\\ step} & \thead{End\\step} & \thead{MAE} & \thead{Size\\(MB)} & \thead{MAE} & \thead{Size\\(MB)} & \thead{MAE} & \thead{Size\\(MB)} \\
\midrule
Dynamic(Best Accuracy) & 0.00 & 0.50 & 0.00 & 60.00	& 8.54 & 22.08 & 8.54 & 21.82 & 8.58 & 6.24\\
\addlinespace
Dynamic(Best model size) & 0.00 & 0.88 & 40.00 & 80.00 & 12.31 & 9.22 & 11.98 & 8.58 & 12.17 & 2.54\\
\midrule
Constant((Best Accuracy) & 0.00 & 0.50 & 60.00 & 80.00 & 8.57 & 35.23 & 8.57 & 35.24 & 8.59 & 7.85\\
\addlinespace
Constant(Best model size) & 0.00 & 0.88 & 60.00 & 80.00 & 11 & 9.21 & 12.39 & 8.41 & 12.51 & 2.59\\
\bottomrule
\end{tabular}
\label{bestModels}
\end{table*}

\begin{figure}
    \centering
    \includegraphics[width=8.5cm]{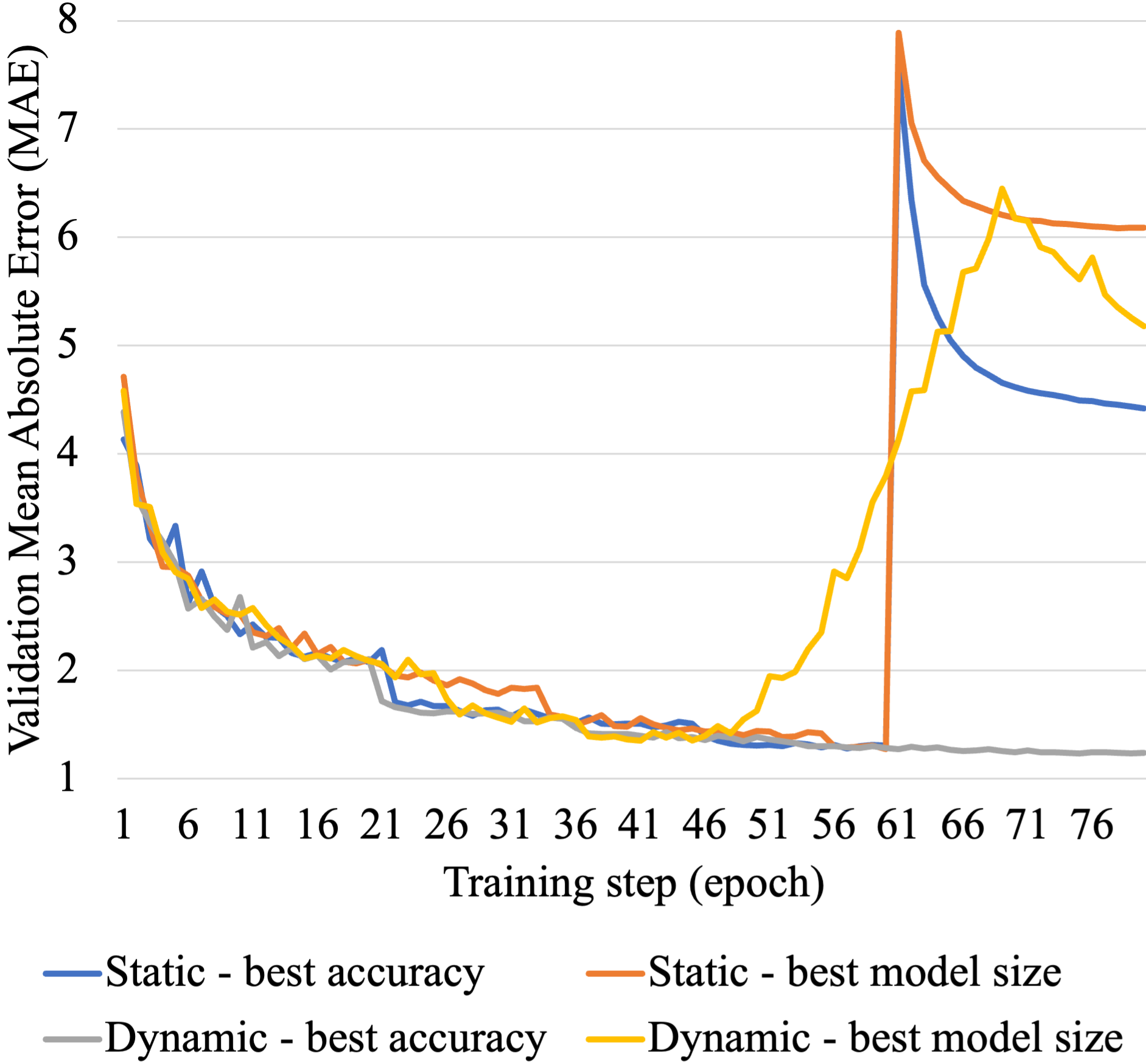}
    \caption{Validation error (MAE)}
    \label{fig:pruneTr}
\end{figure}

According to the Table \ref{bestModels}, we can observe that the model accuracy is better in all three types of models in the dynamic pruned models. The dynamic pruned model is better in 0.03 than the constant pruned model. It is not a significant improvement, but the file size is 13.7MB lower than the constant pruned model. Therefore, with a considerably lower file size, a model with better or at least almost equal accuracy can be obtained by dynamically pruning the model. In the models with the smallest file sizes, the dynamically pruned models still have better accuracy and smaller file sizes than the constant-pruned models. Therefore, we can decide that with dynamic pruning, models can recover the accuracy loss that happens due to pruning better than with constant pruning, and dynamic pruning has pruned more weights than constant pruning, resulting in a model with a smaller file size than constant pruning.

Fig. \ref{fig:pruneTr} shows the variation of validation accuracy over the training steps of the models with the best file size and accuracy. The validation accuracy has suddenly increased at once and rapidly decreased but has stopped at a much higher value than dynamic pruning when constant pruning is applied. But with dynamic pruning, the validation accuracy has increased a bit, has recovered after a few training steps, and has converged to a much lower value than with constant pruning. Also, we can observe that in constant pruning, the ability to recover the damage due to model pruning is lower than in dynamic pruning. This behaviour proves that dynamic pruning can continue the training process without losing the model accuracy and can generate a more accurate model than constant pruning.

\subsection{Post Optimisation Flags: Post-pruning and Quantisation}
\begin{figure}
    \centering
    \includegraphics[width=8.5cm]{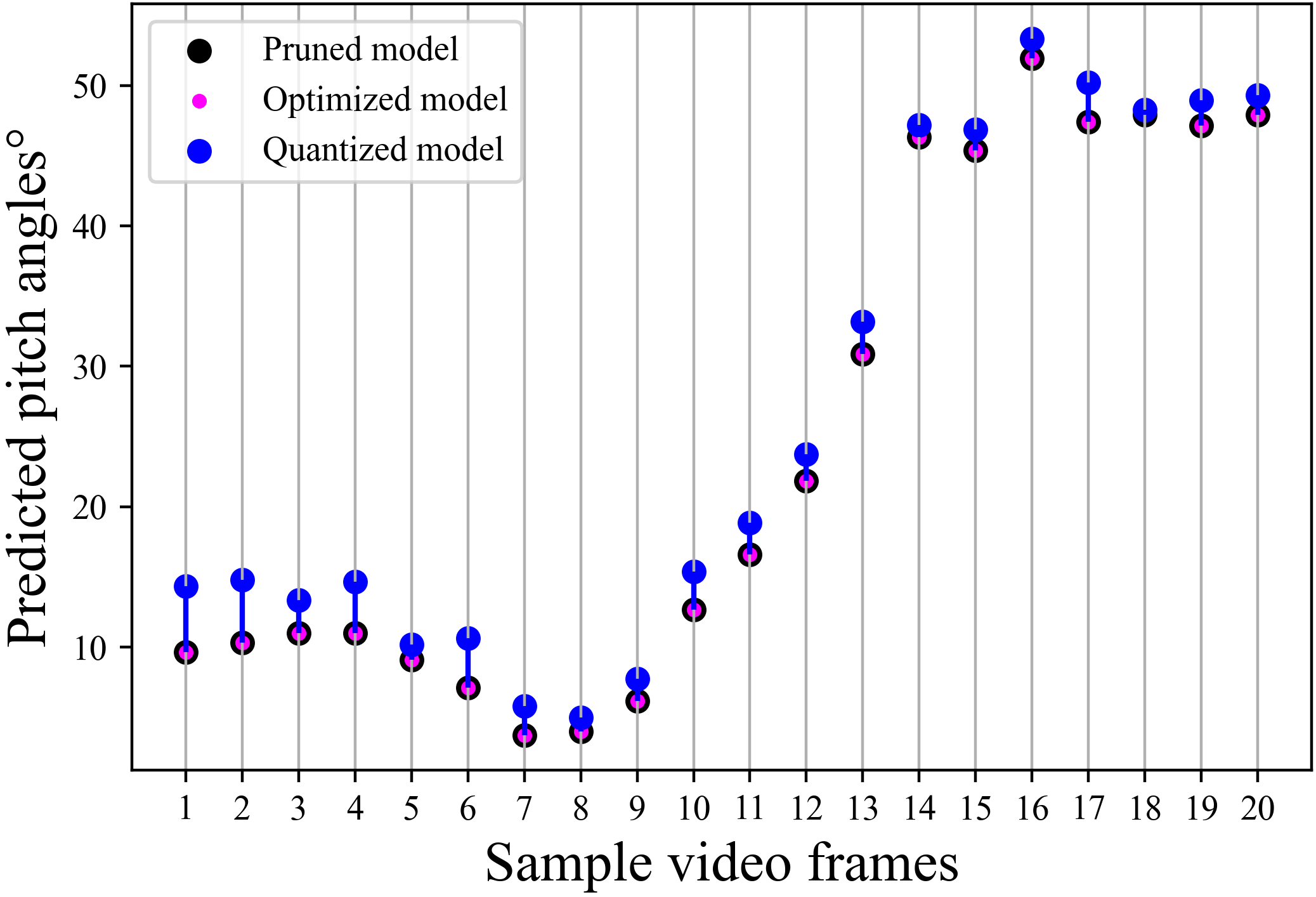}
    \caption{Pitch angle predictions of best accuracy model on sample video frames}
    \label{fig:pitch_pred}
\end{figure}

Post-quantisation can reduce the file size dramatically with a very small loss in accuracy. For example, the model size with the best accuracy can be reduced by 15.84MB with a loss of accuracy of 0.04 by applying post quantisation. Similarly, by observing the graphs described in the previous sections, we can see that the quantisation is very effective in further reducing the model size significantly without losing a considerable amount of accuracy. But Fig. \ref{fig:pitch_pred} shows how much deviation exists between predictions made by the best accuracy model without post-optimisation and the two post-optimised models. Since the post-pruning optimised model has the same accuracy as the pruned model, there's no difference between the pruned model and the optimised model. Therefore, they overlap on each other while the quantized model has a considerable deviation from the pruned model prediction. Although the MAE difference between the pruned model and the post-quantized model is very small (0.04), there's a noticeable deviation between the predictions made by the two models. Therefore, we can conclude that by post-quantisation, although we reduce the file size considerably with a very small accuracy loss, it still impacts the accuracy of the predictions.

\subsection{Non-compressed model sizes}
Without GZip compression, the model sizes remained constant for each type of the model regardless of the pruning parameters in both dynamic and constant pruning, as shown in Table \ref{modelSizes}. Therefore, the best use of pruning is when the model is being transmitted over a network or stored as a compressed file. Only by applying quantisation or post-pruning optimiser, the model size can be reduced further if the model is stored without compression.

\begin{table}[b]
\centering
\caption{Non-compressed model sizes}
\begin{tabular}{l l l l} 
\toprule
\thead{Model\\Type} & \thead{Pruned\\Model(MB)} & \thead{Post Pruned\\Model(MB)} & \thead{Post Quantized\\Model(MB)}\\ 
\cmidrule{0-3}
Dynamic & 36.4MB & 7.4MB & 9.9MB\\
\midrule
Constant & 36.4MB & 7.4MB & 9.9MB\\
\bottomrule
\end{tabular}
\label{modelSizes}
\end{table}

\section{Recommendations}

\begin{figure}
    \centering
    \includegraphics[width=8.5cm]{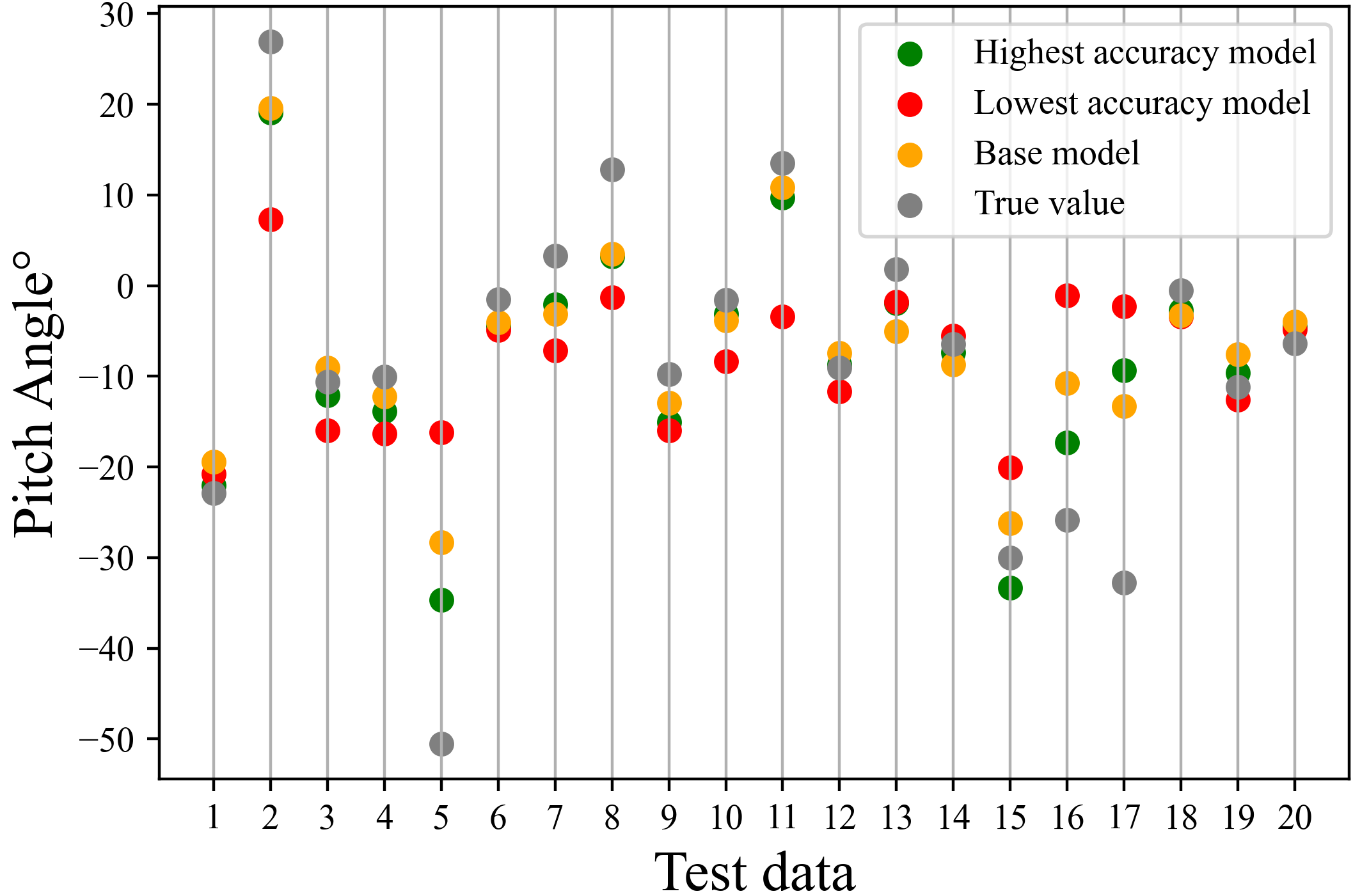}
    \caption{Test data vs Model predictions of best and worst accurate models}
    \label{fig:preidcitons}
\end{figure}

The model accuracy trades off with the model size. At more of the desired sparsity, the model size reduces, but the accuracy of the model also keeps decreasing. Based on the capabilities of the edge device on which the model is going to be deployed, the choice has to be made if the priority has to be given to the model accuracy or to reduce the model size to the lowest possible size with a lower accuracy or reduce the model size up to a moderate level (size less than 50\% of the original model) without reducing the model accuracy or even with better accuracy than the original model. Throughout the experiments, the range of model MAE was 3.77. In order to understand the impact of this variation on the predictions, the pitch angle of 20 consecutive pieces of test data was tested against the pruned models which had the best and worst accuracy. According to Fig. \ref{fig:preidcitons} it can be observed that the model with the best accuracy has a very low distance between the base model and the least accurate model. Since the most accurate model is more accurate than the base model, it has a lower distance to the true value at some points. But at all points, the lowest value model prediction has a huge variation from the true value compared to the other models. The difference between the base model and the best accurate model is (0.09) but still, it can be observed that there's a distance between the predictions of those two models. Therefore, it can be concluded that even a small degradation in model accuracy matters to the accuracy of the model prediction. Based on these observations, the model accuracy vs file size trade-off has to be carefully decided. The primary concern has to be the accuracy, unless the model size is very critical.

We can observe that with post-quantisation we can further reduce the model size with a very small loss in accuracy. With the experimental results of this study for the optimal model pruning output, a range of 0 to 50 dynamic sparsity can be recommended to get a model with good accuracy and over 50\% reduced file size. If the resources in the edge device where the model was to be deployed were more constrained, the final sparsity ($s{_f}$) could be increased to 87.5\%. Based on the findings of this study, the initial sparsity ($s{_i}$) can be recommended to be kept constant at 0. From constant sparsity and dynamic sparsity, dynamic pruning can be recommended since it can generate more accurate models with a lower file size than Constant pruning.

The authors of the pruning module \cite{Prune_not_to_prune} have stated that this technique does not consider the architecture of the model. Since magnitude-based model pruning only looks at the magnitude of the weights, Therefore, regardless of the network architecture, the model pruning technique, along with the observations and conclusions obtained through this study, can be applied to any type of regression model.

\section{Conclusion and Future work}
This work presents an in-depth analysis of model pruning parameter selection in order to obtain a model with a smaller file size without reducing model accuracy. The impact of constant and dynamic pruning and how parameters of them have to be chosen to get an optimal result is discussed in depth throughout this study.

The next step of this study is to combine model pruning with other techniques such as weight pruning, weight clustering, and quantisation and test the performance in different configurations and analyse what the best configuration is to compress models without losing accuracy.

\section{Experimental results and codes}
Experimental results and the model pruning code can be found at: \href{https://github.com/asirigawesha/PruningTests.git}{https://github.com/asirigawesha/PruningTests.git}

\section{Acknowledgement}
 This research was supported by the Accelerating Higher Education Expansion and Development (AHEAD) Operation of the Ministry of Higher Education of Sri Lanka funded by the World Bank (https://ahead.lk/result-area-3/).
 
\bibliographystyle{IEEEtran}
\bibliography{references}
\end{document}